\documentclass[letterpaper, 10 pt, conference]{ieeeconf}  

\IEEEoverridecommandlockouts                              
\usepackage{graphicx} 
\graphicspath{ {./figures/} } 

\usepackage{color}
\usepackage{url}
\usepackage{amsmath} 
\usepackage{amssymb}  
\usepackage{subcaption}
\usepackage{soul}
\usepackage{caption}
\usepackage{array, booktabs, multicol, multirow}
\usepackage[dvipsnames]{xcolor}

\DeclareMathOperator*{\median}{median}

\title{\LARGE \bf
Pneumonia Detection in Chest Radiographs 
}

\author{\parbox{3 in}{\centering The DeepRadiology Team$^{1}$
        \thanks{$^{1}$Contact: Jameson Merkow, \protect\url{jameson@deepradiology.com}}
}
}

\begin{document}

\maketitle
\thispagestyle{empty}
\pagestyle{empty}

\begin{abstract}

In this work, we describe our approach to pneumonia classification and localization in chest radiographs. 
This method uses only \emph{open-source} deep learning object detection and is based on CoupleNet, a fully convolutional network which incorporated global and local features for object detection.
Our approach achieves robustness through critical modifications of the training process and a novel ensembling algorithm which merges bounding boxes from several models.
We tested our detection algorithm tested on a dataset of 3000 chest radiographs as part of the 2018 RSNA Pneumonia Challenge; our solution was recognized as a winning entry in a contest which attracted more than 1400 participants worldwide.

\end{abstract}

\section{Introduction}

Recent advances in computer vision demonstrate that Convolutional Neural Network (CNN) architectures achieve human-level performance on several image-processing tasks, including classification, segmentation, and object detection. As a natural extension of this, many on-going efforts exist to leverage its capabilities in medical imaging. 
Clinicians utilize such algorithms for assistance in anomaly detection in a variety of medical imaging modalities. 
For example, CNNs achieve performance on par with board-certified dermatologists for the classification of skin cancer on biopsy-proven clinical images \cite{thurn_nature_2017} and board-certified radiologists for pathology detection in computerized tomography head scans \cite{vedaldi_arxiv_2017}.

The next frontier for AI in medicine is \emph{anomaly localization} in medical imaging. \emph{Anomaly localization} refers to predicting anomalies as well as their boundaries. Currently, few CNN-based anomaly localization algorithms exist because training these algorithms requires a large number of professionally curated images for each disease type and modality of measurement. 
However, the RSNA Pneumonia Detection Challenge \cite{c40} provided such a dataset and it became a testing ground for anomaly localization algorithms, attracting over $1400$ participating teams.

Here, we present our winning solution to the RSNA Pneumonia Detection Challenge. 
We will discuss our training, inference and evaluation methodologies that were based on an ensemble of the CoupleNet \cite{c13} object detection architecture.

\section{Background}

\subsection{Relevant Work}

P. Viola and M. Jones \cite{viola_jones} introduced the first efficient object detection (face recognition) algorithm in which they introduced a novel image representation called \emph{integral image}. 
Half a decade later, N. Dalal and B. Triggs outperformed existing feature sets by developing Histogram of Oriented Gradient (HOG) for pedestrian detection \cite{hog}. 
After another half a decade, Felzenszwalb et al. \cite{c11} developed the Deformable Parts Model (DPM) that uses mixture model of deformable parts to recognize objects.

More recently, in 2012, Kriszhevsky et al. \cite{c3} used Convolutional Neural Networks (CNN) to outperform every other algorithm on ImageNet dataset. 
Significant advances in this field has been made since then \cite{c4, c5, c6, c7, c8, c9, c10}.  
Since Sermanet et al. \cite{c20} proposed an integrated framework using deep neural network for object detection in 2014, many others have been developed since, for e.g. Fast-RCNN \cite{c14}, Faster-RCNN \cite{c15}.

Two predominant deep neural network based object detection approaches exist today. The first, single-shot detectors, pose object detection as a regression problem and perform object detection as well as classification in a single forward pass of the network. 
This method has been used successfully in a numbers of approaches, such as 
Single Shot MultiBox Detector (SSD) \cite{c37}, You Only Look Once (YOLO) family, \cite{c33,c34,c35} and Feature Pyramid Network (FPN) \cite{c36}.
The second method, based on region proposals, use a multi-stage pipeline that simultaneously learns to propose accurate bounding boxes and classify them.
Since, our winning entry uses a region proposal-based detector, we will keep our description focused to region proposal based methods.

Region-based Convolutional Neural Networks (R-CNN) \cite{c12} use a multi-stage approach to extract features from each region proposal for classification and simultaneously train a regression model to correct predicted bounding-box from learned offsets. 
However, it is computationally expensive because of its multi-stage approach without shared computation. 
To combat this drawback, Girshick proposed two improvements Fast R-CNN and Faster R-CNN which unified the models to take advantage of shared computation \cite{c14}, and further improved speed by combining models \cite{c15}. 
R-FCN improved this further by removing several fully-connected layers per-RoI (region of interest) and replacing them with position-sensitive RoI-pooling \cite{c16}.

The authors of \cite{c13} added a secondary branch to R-FCN which processed global features. 
The resulting architecture, CoupleNet, merges features from \textit{global} and \textit{local} branches. 
The \textit{global} branch learns features from a larger area around each RoI and adds additional per-RoI convolutional layers while the \textit{local} branch concentrates local information using position-sensitive score maps similar to those in R-FCN.

\begin{figure*}[htp]
  \centering
  \begin{minipage}[b]{.95\textwidth}
  \centering
  \includegraphics[width=\textwidth]{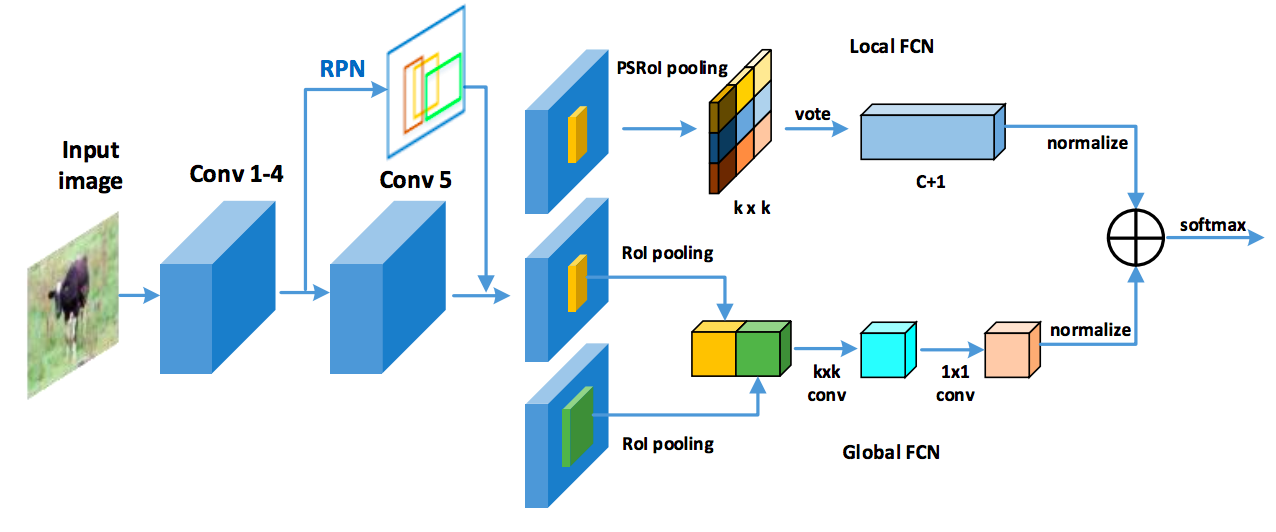}
  \caption{Illustration of CoupleNet architecture \cite{c13}. Images are passed through a base network which feeds into a Region Proposal Network (RPN) to generate proposals. Each proposal is passed to two branches. The first uses Position-sensitive RoI (PSRoI) pooling to capture local information. The second branch extracts global context information. The output of these two branches is merged to generate predictions based on both local and global information.}
  \label{fig:couple-net}
   \end{minipage}
\end{figure*}

Medical imaging technologies started  these advances in detection algorithms as early as mid 1990's for lung nodules localization in chest X-rays \cite{c25}.
Since then there have been a number of works that utilize CNNs for detection.
Recently, which is composed of CT scans was released. Yan et al. 
\cite{c45} experimented on the DeepLesion dataset \cite{c46} with their 3D based modified version of R-FCN to effectively detect lesions.
Ravishankar et al. \cite{c29} combined classical features with CNN pre-trained on ImageNet data to detect kidney in ultrasound (US) images with higher degree of performance.
Wang et al. \cite{c40} presented chest X-ray dataset and used a heatmap based approach to generate bounding boxes in a weakly supervised manner.
Hwang et al. \cite{c42} evaluated a CNN framework which simultaneously optimizes classification and localization networks on two public Chest X-rays and Mammograms datasets.
Li et al. \cite{c43} combined activated patches obtained from ResNet architecture feature maps to learn localization information on chest X-rays.
Pesce et al. \cite{c44} exploited saliency maps obtained from CNN and combined them with recurrent network and reinforcement learning to detect pulmonary nodules in chest X-rays.

\subsection{RSNA Pneumonia Detection Challenge}
In this section, we describe the details of the RSNA Pneumonia Detection Challenge.
As typical of competitions of this nature, it was presented in two stages.
During the first stage, the competitors were asked to develop a training and inference algorithm based on training and testing data. 
The first stage attracted over 1400 particants.
In the next stage, each teams chosen methodology was evaluated with new training and testing set.

The RSNA Pneumonia Detection Challenge dataset is a subset of 30,000 exams taken from the NIH CXR14 dataset \cite{c24}. 
From the 30,000 selected exams, 15,000 exams had positive findings for pneumonia or similar pathologies such as consolidation and infiltrate. 
The 15,000 negative exams were taken from two groups: 7,500 exams had no findings and 7,500 exams had pathologies unrelated to pneumonia. 
Six board-certified radiologists annotated all positive samples in the dataset with bounding boxes using a commercial annotation system. 
Three total radiologists, one of which was a radiologist from the annotators and two new radiologist from the Society of Thoracic Radiology (STR), confirmed 3,000 images for the stage I and II test sets. 
More details on the data collection can be found here \cite{c21}.

Stage I consisted of 25,684 training images and 1,000 testing images. For stage II, the 1000 testing samples where added to training to form dataset of 26,684 training images and a new set of 3,000 radiographs were introduced for test.
During both stages, the competition provided annotations for only the training set.

During the competition, participants submitted up to 5 predictions per day to the competition server which were scored based on Eq. \ref{eqn:iou_dataset}. 
During stage I, the score reflected results from the entire test set, however, during stage II, participants received feedback on only $1\%$ of the stage II test set. Upon the conclusion of the competition, scores for the remaining $99\%$ were released, generating the final standings. 

\begin{figure}[htp]
    \centering
    \begin{minipage}[b]{.45\textwidth}
    \centering
    \includegraphics[width=\textwidth]{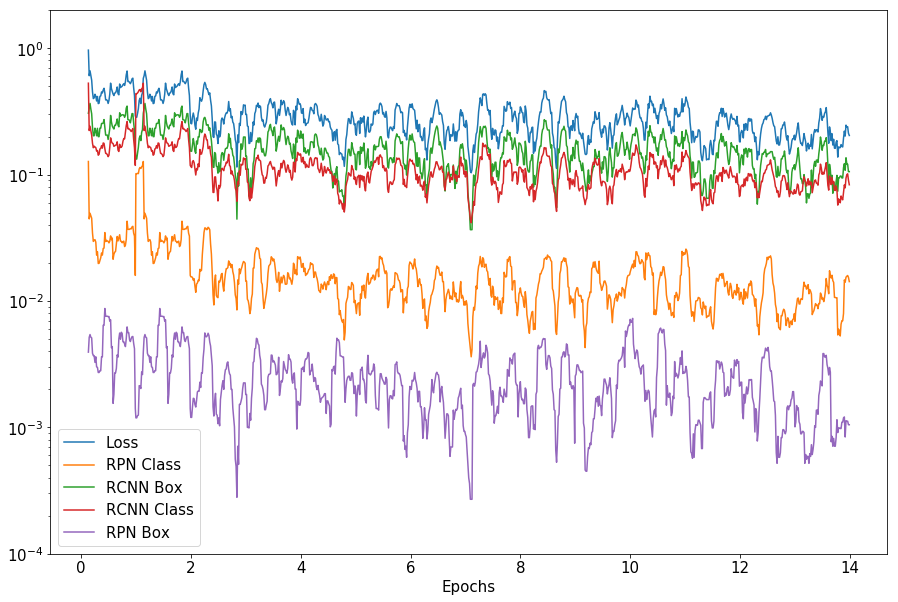}
    \caption{Graph of typical training losses. Blue shows loss for the overall network, orange and purple depict RPN class loss and RPN box loss, respectively. Red and green show RCNN class loss and RCNN box loss.}
    \label{fig:train_log}
    \end{minipage}
\end{figure}

\section{Algorithm}

CoupleNet \cite{c13} formed the basis for our detection algorithm. While we reviewed several \textit{open-source} object detection architectures, from our experiments, we found that CoupleNet produced the strongest results. In our implementation, we used the ResNet-152 \cite{c5} pre-trained on ImageNet as our base network. We also used RoI Cropping \cite{c17} for the global branch rather than RoI Pooling. 

We trained our models end-to-end using multi-task loss \cite{c15} and evaluated hyper-parameters on a validation set consisting of a $10\%$ stratified sample of the training set. 
During training, we used stochastic gradient descent (SGD) with an initial learning rate of $0.001$. 
This learning rate was dropped once at $10$ epochs by $90\%$ and our models trained for a total of $14$ epochs. 
We set the foreground $IoU$ threshold to $0.30$ so the RoIs contained small portions of the ground-truth for both bounding box coordinates and class scores.
This is inline with the competition metric, which scored boxes at lower ground-truth $IoU$ thresholds.
During training we augmented the data with random horizontal flipping as well as random re-scaling within batches.

Final predictions were generated by an ensemble of four of models trained on the entirety of the training set.
Each model produces unique predictions, of these, we only considered bounding boxes with a confidence threshold of $0.50$ or above, which were ensembled as follows.
First, bounding boxes from all models were clustered such that those with an $IoU \geq 0.25$ were grouped.
Next, for each bounding box group, we combined the confidence scores using the following equation:
\begin{equation}
\label{eqn:score-ens}
    \hat{s}_m = \frac{1}{N} \sum_{i} s_{m,i}
\end{equation}
where $s_{m,i}$ is the confidence score for the $i^{\text{th}}$ bounding box for the $m^{\text{th}}$ group, $N$ represents a scaling factor (set to $4$), and $\hat{s}_m$ is the ensembled confidence score for the $m^{\text{th}}$ group. 
Each corner of the ensembled bounding box for the $m^{\text{th}}$ group is found by: 
\begin{equation}
\label{eqn:box-ens}
    \hat{C}_{ml} = \median\left\{\mathbf{C}_{ml}\right\} + \alpha \cdot \sigma_{ml}
\end{equation}
where $\mathbf{C}_{ml}$ represents the set of pixel locations of the corner $l$ (top-left, top-right, bottom-left, or bottom-right) for each $m^{\text{th}}$ group of bounding boxes, $\sigma_{ml}$  represents the standard deviation of $\mathbf{C}_{ml}$, $\alpha$ represents the scaling factor (set to $0.1$), and $\hat{C}_{ml}$ represents the ensembled bounding box pixel location for corner $l$. 
Any ensembled bounding box with a confidence score less than $0.25$ were excluded from the final bounding box predictions.

\section{Results}
\begin{figure*}[htp]
    \centering
    \begin{minipage}[b]{.95\textwidth}
    \centering
    \begin{subfigure}{\textwidth}
        \centering
        \includegraphics[width=\textwidth]{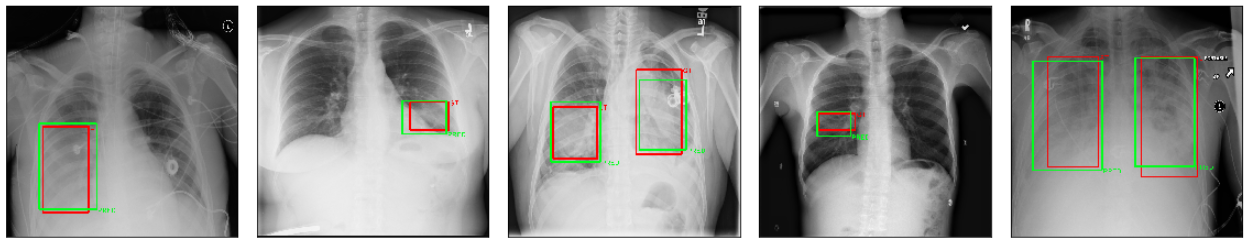}
    \end{subfigure}
    
    \begin{subfigure}{\textwidth}
        \centering
        \includegraphics[width=\textwidth]{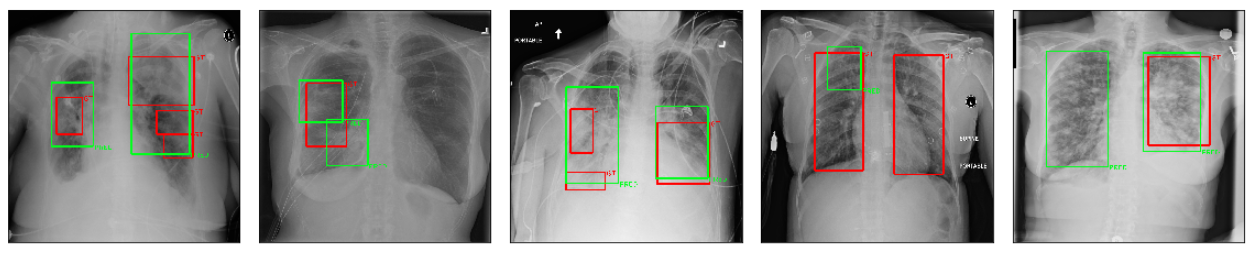}
    \end{subfigure}
    \caption{Example results from our stage I detector on the stage I test set. The top row depicts successful predictions and the bottom row shows errors. Algorithm predictions and ground truth shown as green and red overlays, respectively.}
    \label{fig:model_detection}
    \end{minipage}
\end{figure*}

\subsection{Metrics}

The competition evaluation metric was determined using the mean of the intersection over union ($IoU$) of matching prediction and ground-truth boxes at several matching thresholds. The matching threshold is based on the $IoU$ of the areas of the ground-truth and prediction boxes. The following formula computes the $IoU$ of the areas between a ground-truth and prediction box:
\begin{equation}
\label{eqn:iou}
IoU_{area}(B_{true}, B_{predicted}) = \frac{B_{true} \cap B_{predicted}}{B_{true} \cup B_{predicted}}
\end{equation}

The $IoU$ between prediction and ground-truth boxes is then thresholded using $8$ thresholds values, ranging from $0.4$ to $0.75$, to find "hits" and "misses". 
Matches between prediction boxes and ground-truth boxes is strictly one-to-one and are matched in descending order of the predictions confidence score.

At each threshold value, the number of true positives ($TP$), false negatives ($FN$), and false positives ($FP$) are determined and the score coefficient ($C$) over the results are found for a specified threshold $t$:
\begin{equation}
\label{eqn:ppv}
C(t) = \frac{TP(t)}{TP(t)+FP(t)+FN(t)}
\end{equation}
Next, we find the mean score ($C$) over all threshold values in each image:
\begin{equation}
\label{eqn:image_iou}
C_{i} = \frac{1}{|T|} \sum_t^T C(t)
\end{equation}
where $C_{i}$ represents the mean score coefficient on $i^{\text{th}}$ image and $T$ is the set of specified thresholds such that $t \in T$.
Finally, we find the mean score for the dataset with:
\begin{equation}
\label{eqn:iou_dataset}
mC_{dataset} = \frac{1}{|I|} \sum_i^I C_{i}
\end{equation}
where $mC_{dataset}$ represents the score for the entire dataset, $I$ represents images in the dataset that contain either a ground truth or predicted bounding box.

\subsection{Findings}

Fig. \ref{fig:train_log} depicts a typical training log for our models. In Fig. \ref{fig:model_detection}, we provide examples of bounding box predictions from the stage I test set, in which we had ground-truth annotations after stage II commenced; the top row show successful predictions and the bottom row depicts discrepancies between ground-truth and prediction boxes. 
Table \ref{tab:map} shows our results using our ensembling algorithm as well as a single model average from that ensemble where we see strong improvement from using the ensemble method over the average model score in both stages.
\begin{table}[h]
    \renewcommand{\arraystretch}{1.5}
    \centering
    \caption{RSNA Pneumonia  Detection  Challenge Results.}
    \begin{tabular}{m{8em}|m{15mm}m{15mm}}
        \  & Stage I & Stage II \\ \hline
        Model Average & $0.2160$ & $0.2167$ \\ \hline
        Ensemble & $0.2180$ & $0.2310$ \\
        \bottomrule
        \end{tabular}
    \label{tab:map}

\end{table}

\section{Conclusion}
Three factors played an important role in our final competition score which resulted in a winning solution in the RSNA Pneumonia Detection Challenge. 
First, choosing an architecture with global and local context, such as CoupleNet, provided extra context for generating accurate results. 
Second, using sensible foreground and background proposal thresholds while training tuned our network to perform well in the competition.
Finally, as shown in Fig. \ref{tab:map}, our ensemble algorithm significantly boosted prediction consistency and improved accuracy.
We continue to work on this and in other areas of medical imaging, applying cutting-edge computer vision and deep learning technology to deliver unparalleled automation and aid for health-care services.

\section*{Acknowledgements}
We would like to thank the RSNA and Kaggle staff for contributing their time and expertise to put on this competition.  We had a great time participating and we were excited to see the strong turn out.




\end{document}